\documentclass[prl,twocolumn,amsmath,nobibnotes]{revtex4-1}

\usepackage[utf8]{inputenc}
\usepackage{graphicx, floatrow}
\usepackage[caption=false]{subfig}
\usepackage{amssymb, amsfonts, amsmath, dsfont}
\usepackage{comment}
\usepackage{multirow}
\usepackage{makecell}
\usepackage[T1]{fontenc}
\usepackage{floatrow}
\usepackage[table]{xcolor}
\usepackage{letltxmacro}
\usepackage[section]{placeins}

\usepackage[]{algorithm2e}
\usepackage{hyperref}

\LetLtxMacro{\originaleqref}{\eqref}
\renewcommand{\eqref}[1]{\originaleqref{eq:#1}}

\def \del{\partial}    

\def \>{\rangle} 
\def \<{\langle}

\def\bv{\textbf{v}}
\def\bh{\textbf{h}}

\def\bb{\textbf{b}}

\def\be{\begin{equation}} 
\def\ee{\end{equation}} 
\newcommand \bea {\begin{eqnarray}} 
\newcommand \eea {\end{eqnarray}} 
 
\newcommand{\nn} {\nonumber}



\begin{document}

\title{Can RBMs be trained with zero step contrastive divergence?}

\author{Charles K. Fisher}
\email{drckf@unlearn.ai}
\affiliation{Unlearn.AI, Inc., 75 Hawthorne St. Ste 560, San Francisco, CA 94105}

\date{\today}

\begin{abstract} 
Restricted Boltzmann Machines (RBMs) are probabilistic generative models that can be trained by maximum likelihood in principle, but are usually trained by an approximate algorithm called Contrastive Divergence (CD) in practice. In general, a CD-k algorithm estimates an average with respect to the model distribution using a sample obtained from a k-step Markov Chain Monte Carlo Algorithm (e.g., block Gibbs sampling) starting from some initial configuration. Choices of k typically vary from 1 to 100. This technical report explores if it's possible to leverage a simple approximate sampling algorithm with a modified version of CD in order to train an RBM with k=0. As usual, the method is illustrated on MNIST. 
\end{abstract} 

\maketitle

Restricted Boltzmann Machines (RBMs) are probabilistic generative models composed of two layers of neurons (called the visible and hidden layers) with undirected connections between them \cite{ackley1985learning}. Because RBMs are energy based models, it's relatively easy to derive the gradient of the log-likelihood function. Unfortunately, computing the gradient requires one to calculate averages with respect to the model distribution, which are typically intractable. As a result, RBMs are usually trained using approximate methods in which the required averages are estimated using Markov Chain Monte Carlo (MCMC) methods such as block Gibbs sampling \cite{hinton2006training, tieleman2008training, hinton2010practical}. 

Liao et al \cite{liao2022gaussian} recently reported that a few changes to the typical procedure for training RBMs significantly improved the performance of RBMs with Gaussian visible units on image generation tasks. First, Liao et al replace the exact Gibbs sampling step of the visible units with an approximate sampling method based on Langevin dynamics. Second, they initialize the samples used to compute the negative phase of the gradient with an isotropic Gaussian noise such that sampling the visible units can be viewed as type of denoising algorithm. 

A gradient step of an RBM aims to decrease the energy of the observed samples (i.e., the positive phase) and increase the energy of the samples drawn from the model (i.e., the negative phase). Typically, one raises the energy of samples drawn from an approximation of the model distribution using something like k-step Contrastive Divergence (CD-k) with a large number of MCMC steps, but does this approximation actually need to be good? To explore this question, this technical report asks whether it's possible to train an RBM with a version of CD-0.

To be concrete, I will focus on RBMs with discrete visible and hidden units. I use the physics convention with Ising-like neurons that are $\pm1$ rather than Bernoulli units that are $(0,1)$, but the two model types are theoretically the same as they are related by a simple linear change of variables. The joint energy function of this model is,
\be
U(\bv, \bh) = - \bb^T (\bv - \boldsymbol{\mu}) - (\bv - \boldsymbol{\mu})^T W \bh \, .
\ee
Note that I have centered the visible units about their empirical mean $\boldsymbol{\mu} = \< \bv \>_{d}$, and I have not included a separate bias on the hidden units; though, that can easily be done. Neglecting the explicit bias on the hidden units effectively sets the bias to $W^T \boldsymbol{\mu}$, which works fine on binary MNIST.

The probability distribution of the visible units is defined in terms of the energy function by,
\be
p(\bv) = \frac{ \sum_{H} e^{-U(\bv, \bh)} }{ \sum_V \sum_H e^{-U(\bv, \bh)} },
\ee
in which I use the notation $\sum_V$ and $\sum_H$ to denote sums over all allowed values of the visible and hidden units. Thus, the negative log-likelihood is,
\be
\mathcal{L}(\bb, W) = -\< \log p(\bv) \>_{d} \, .
\ee
As usual, one fits the model by attempting to minimize the negative log-likelihood.

Stochastic gradient descent, in its various forms, is great at minimizing negative log-likelihoods. In order to use it, however, one needs expressions for the gradient of the negative log-likelihood with respect to the model parameters. These can be computed as,
\begin{align}
\frac{\del \mathcal{L}}{\del \bb} &= \< \bv \>_{d}  - \< \bv \>_{m} \, , \\ 
\frac{\del \mathcal{L}}{\del W} &= \< \bv \tanh( W (\bv - \boldsymbol{\mu}))^T \>_{d} \nn \\
&- \< \bv \tanh( W(\bv - \boldsymbol{\mu}) )^T \>_{m} \, .
\end{align}
The expressions for the gradients contain two types of averages, one taken over the samples in the training dataset and one taken with respect to the distribution defined by the model itself. The former average is easy to calculate, but the latter is intractable. 

In order to calculate averages with respect to the model distribution, i.e. $\< f(\bv) \>_{m}$, one needs to draw samples from the marginal distribution $p(\bv)$. Typically, this is done through block Gibbs sampling \cite{hinton2010practical}. To perform block Gibbs sampling, we start with an initial hidden state $\bh_0$ and then iteratively draw from $p(\bv_i | \bh_{i-1})$ and $p(\bh_i | \bv_i)$ for $i = 1, ..., k$. Each of these conditional distributions is easy to sample from due to the bipartite structure of the interaction graph in an RBM. The problem is that block Gibbs sampling tends to mix slowly, so that many iterations are required in order to generate reasonable samples from the distribution. However, if one could just sample $\bh_0 \sim p(\bh)$ from the marginal distribution of the hidden states then one could get a sample of the visible states with a single draw from $p(\bv | \bh_0)$. 

One can rewrite the conditional distribution $p(\bh | \bv)$ as $p(\bh | \boldsymbol{\phi}) = Z^{-1} e^{-\boldsymbol{\phi}^T \bh}$ in which $\boldsymbol{\phi} = W^T (\bv - \boldsymbol{\mu})$ is a field acting on the hidden units that is induced by the visible units. If the dimension of the visible space is large, then $\phi_i = \sum_j W_{ji} (v_j - \mu_j)$ is a weighted sum of random variables and (under some conditions) is approximately Gaussian. Thus, if $\boldsymbol{\mu}$ and $\Sigma$ are the empirical mean and covariance matrix of the visible units, then $\boldsymbol{\phi} \sim \mathcal{N}(0, W^T \Sigma W)$ is approximately the distribution of the fields under the data distribution. That is,
\be
p(\bv) \approx \sum_{H} \int d \boldsymbol{\phi} p(\bv | \bh) p(\bh | \boldsymbol{\phi}) p(\boldsymbol{\phi}) \, ,
\ee
is way to approximate the marginal distribution of the visible units in an RBM using a stochastic directed neural network. 

This leads to a simple algorithm for generating approximate samples from an RBM using a single backward pass, which we call ``belief generation'' due to the resemblance to a deep belief network. 
\begin{itemize}
\item{Step 1:} Draw $\boldsymbol{\phi} \sim \mathcal{N}(0, W^T \Sigma W)$. 
\item{Step 2:} Draw $\bh \sim Z^{-1} e^{-\boldsymbol{\phi}^T \bh}$. 
\item{Step 3:} Draw $\bv \sim p(\bv | \bh)$ and use $\bv$ as a sample from the model distribution. 
\end{itemize}
Note that this can be done without having to compute $W^T \Sigma W$ every time. Instead, store a matrix $Q$ that is a square root of $\Sigma$ and then set $\boldsymbol{\phi} = W^TQz$ where $z$ is a vector of standard normal random variables. In practice, a few Gibbs sampler updates need to be performed starting from the resulting sample in order to obtain good samples, but maybe good samples aren't required for training?

To address this question, I trained an RBM on binary MNIST using CD-0. The RBM had 512 hidden units, and was trained for 300 epochs with a batch size of 1024 and a constant learning rate of $1e^{-3}$ using the ADAM optimizer \cite{kingma2014adam}. The weights were randomly initialized from a Gaussian distribution with standard deviation 0.1, and no weight decay was used.

\begin{figure}[t!]
\includegraphics[width=\columnwidth]{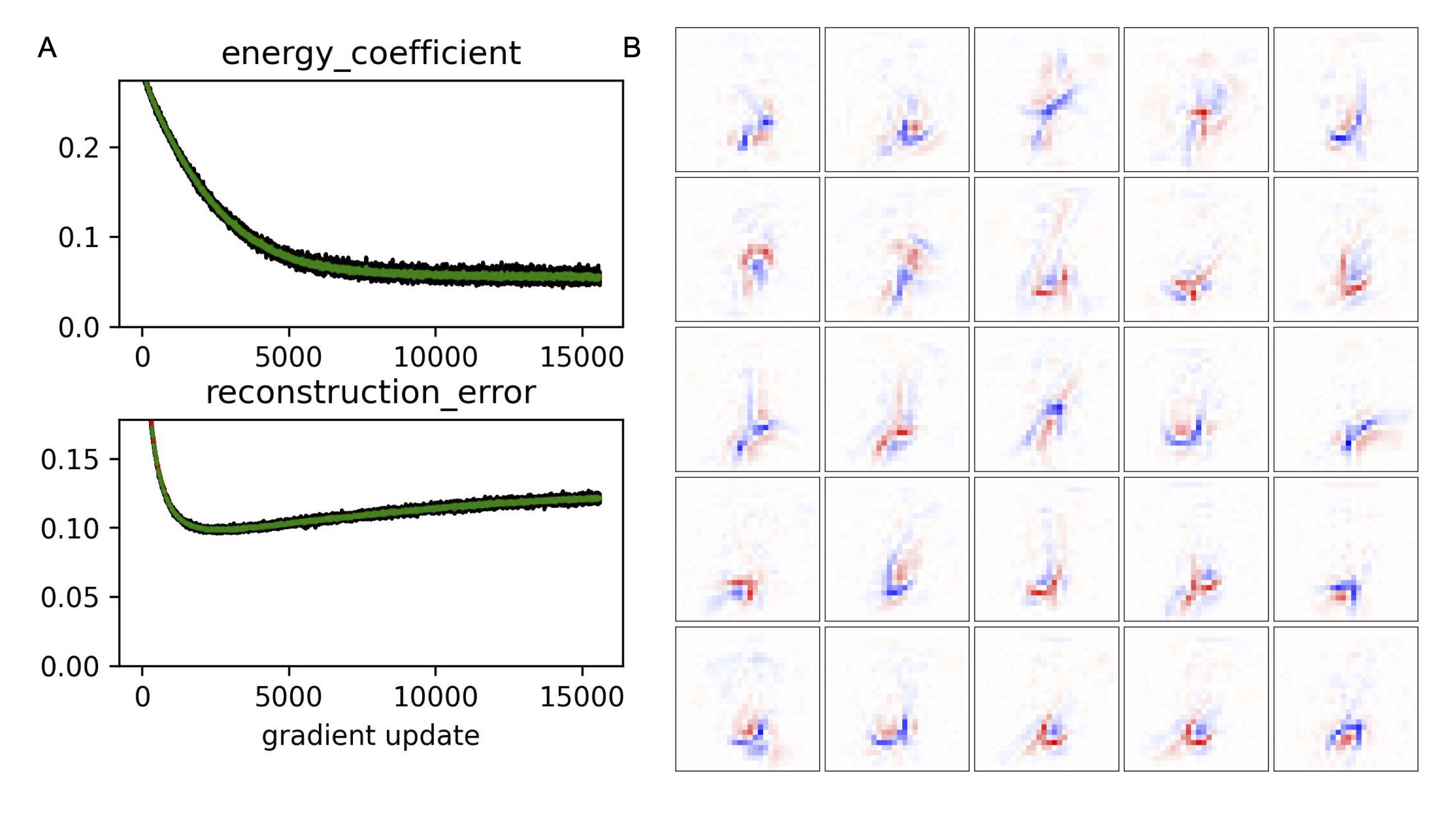}
\caption{\textbf{Training an RBM with CD-0 on MNIST.} A. Training dynamics. B. A random sample of learned weights at the end of training. 
\label{fig:cd-0-training}}
\end{figure}

\begin{figure}[b!]
\includegraphics[width=\columnwidth]{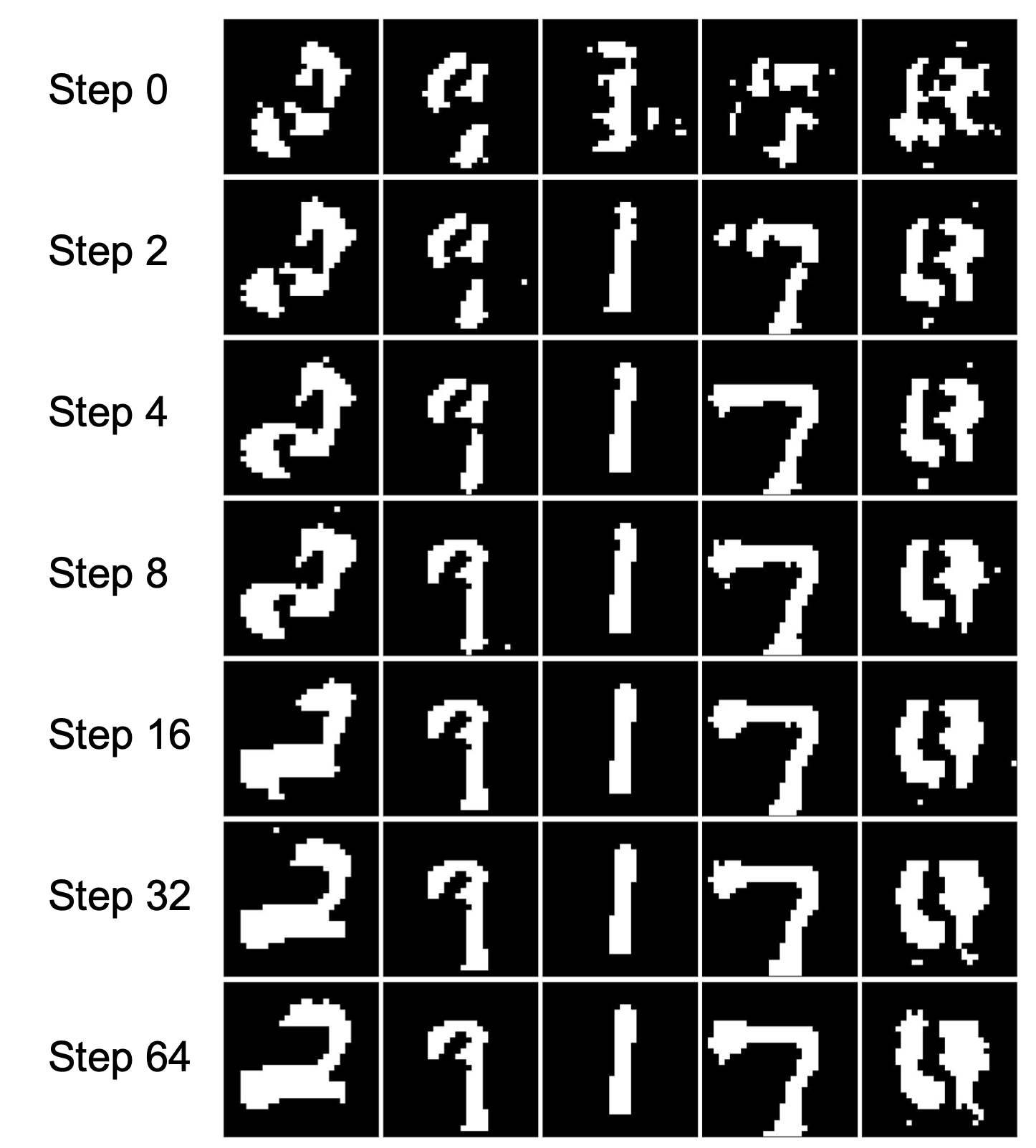}
\caption{\textbf{Samples from an RBM trained by CD-0.} The initial row (Step 0) presents samples directly obtained from the approximate ssampling algorithm, and each subsequent row illustrates how those samples evolve through block Gibbs sampling.
\label{fig:cd-0-samples}}
\end{figure}

\begin{figure*}[t!]
\includegraphics[width=\columnwidth]{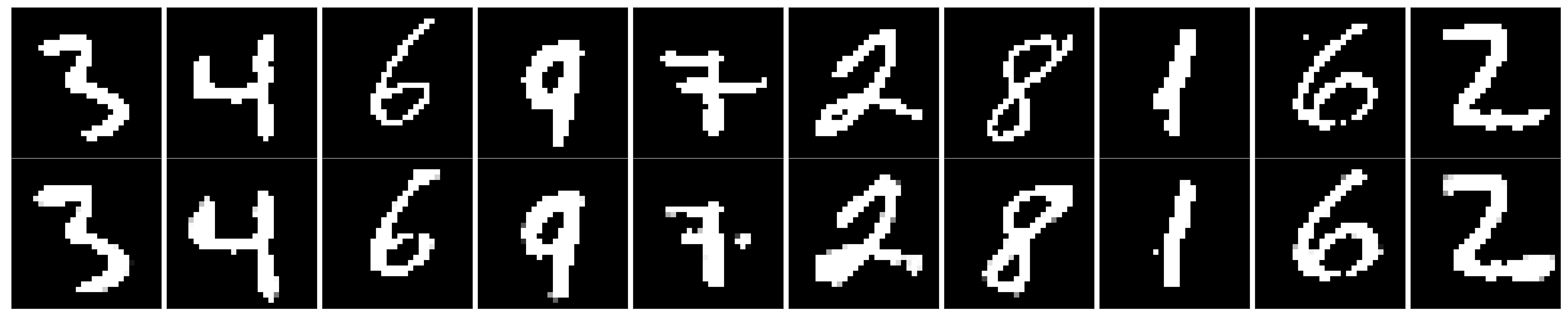}
\caption{\textbf{Reconstructions using an RBM trained by CD-0.} Original on top, reconstruction on bottom.
\label{fig:cd-0-recon}}
\end{figure*}

Figure \ref{fig:cd-0-training}A illustrates the training dynamics using two metrics. The first is a normalized statistical distance between observed and sampled images called the energy coefficient \cite{szekely2003statistics}, 
\be
\frac{2 \< \< ||v-v'|| \>_{d} \>_{m} - \< \< ||v-v'|| \>_{d} \>_{d} - \< \< ||v-v'|| \>_{m} \>_{m}}{2 \< \< ||v-v'|| \>_{d} \>_{m}} \nn \,
\ee
and the second is the 1-step reconstruction error. Both metrics decrease during training as the model learns reasonable localized features (Figure \ref{fig:cd-0-training}B).

Next, Figures \ref{fig:cd-0-samples} shows samples drawn from the model using different numbers of alternating block Gibbs sampling steps. The images in the first row, labeled Step 0, are obtained directly from the belief generation sampling algorithm. These initial samples are updated with different numbers of Gibbs sampling steps, as shown in the subsequent rows. 

The samples obtained from the belief generation sampling algorithm are clearly very noisy, but they are in a sense close to realistic handwritten digits. One can view the Gibbs sampling steps as image denoising, much like accessing a memory in a Hopfield network \cite{hopfield1982neural, mehta2019high}. By initializing CD chains to noisy images, the gradient updates decrease the energy near the observed data points, and increase the energy near the noisy digits to create a basin of attraction so that Gibbs sampling updates move noisy samples closer to realistic samples. 

The samples drawn from the CD-0 trained RBM aren't of particularly high quality -- they are okay, but not great -- but the reconstructions shown in Figure \ref{fig:cd-0-recon} are a different story. The CD-0 trained RBM learns to create faithful reconstructions because the observed samples are encoded in deep minima on the energy landscape. 

One difficulties with RBMs is that the energy function pulls double duty; the topography of the energy function encodes patterns as basins of attraction, but the same topography also determines the rate of mixing during MCMC sampling. The model learns the patterns better as the basins of attraction get deeper, but this leads to slow hopping between basins. As a result, MCMC chains need to be extremely long to sample the whole distribution. Belief generation gets around the need to run long MCMC chains to obtain samples from the RBM. Instead, one draws a batch of initial samples that are close to the basins of attraction, then uses Gibbs sampling updates to move towards the energy minima. 

As the model trains, the 1-step reconstruction error of the observed samples decreases substantially. However, the reconstruction error is not uniform across sample space. Rather, it decreases with each Gibbs sampling step (Table \ref{table:1}) starting from an initial state created with belief generation. Much like a diffusion model, the RBM is iteratively denoising an initial image as it moves down the basin of attraction.

This technical report recasts the process of sampling from an RBM using long Markov Chains into a composition of a simple approximate sampling algorithm followed by a short Markov Chain. This approach may lead to a substantial speedup in training, enabling RBMs to scale to previously intractable problems.

\begin{table}[h!]
\begin{center}
\begin{tabular}{ |c||c|c|c|c|c|c| } 
 \hline
  & Step 0  & Step 2 & Step 4 & Step 8 & Step 16 & Step 32\\ 
\hline\hline
 CD-0 & 0.35 & 0.27 & 0.23 & 0.21 & 0.20 & 0.18 \\ 
 \hline
\end{tabular}
\caption{One-step reconstruction error as a function of the number of Gibbs sampling steps away from the initial state.}
\label{table:1}
\end{center}
\end{table}


\end{document}